\documentclass[10pt, conference]{IEEEtran}
\IEEEoverridecommandlockouts
\usepackage{cite}
\usepackage{amsmath,amssymb,amsfonts}
\usepackage{algorithmic}
\usepackage{graphicx}
\usepackage{textcomp}
\usepackage{xcolor}
\def\BibTeX{{\rm B\kern-.05em{\sc i\kern-.025em b}\kern-.08em
    T\kern-.1667em\lower.7ex\hbox{E}\kern-.125emX}}

\usepackage[binary-units]{siunitx}
\usepackage{comment}
\usepackage{enumitem}
\usepackage{booktabs} % nicer table
\usepackage{multirow} % table
\usepackage[font=small,labelfont=bf]{caption} 
\usepackage[hyphens]{url} %new line on url footnotes

\usepackage[left=1.55cm,top=1.35cm,right=1.55cm,bottom=2.5cm,nofoot]{geometry}
\newcommand{\squeezeup}{\vspace{-1.5mm}}

\usepackage[compact]{titlesec}
\titlespacing{\subsection}{0pt}{*1}{*0}
\titlespacing{\subsubsection}{0pt}{*0}{*0}

\title{Low-Power License Plate Detection and Recognition on a RISC-V Multi-Core MCU-based Vision System}

\definecolor{somegray}{rgb}{0.5, 0.5, 0.5}
\newcommand{\darkgrayed}[1]{\textcolor{somegray}{#1}}
\makeatletter
\newcommand*\titleheader[1]{\gdef\@titleheader{#1}}
\AtBeginDocument{%
  \let\st@red@title\@title
  \def\@title{%
    \vskip-1em
    \bgroup\normalfont\large\centering\@titleheader\par\egroup
    \vskip0.07em\st@red@title}
}

\makeatother
\titleheader{\darkgrayed{This paper has been accepted for publication in the IEEE International Symposium on Circuits and Systems (ISCAS) \copyright 2021 IEEE.}}
\author{
\IEEEauthorblockN{Lorenzo Lamberti\IEEEauthorrefmark{1}, 
Manuele Rusci \IEEEauthorrefmark{1}\IEEEauthorrefmark{3}, 
Marco Fariselli\IEEEauthorrefmark{3}, 
Francesco Paci\IEEEauthorrefmark{3}, 
Luca Benini\IEEEauthorrefmark{1}\IEEEauthorrefmark{2}
}
\IEEEauthorblockA{
\IEEEauthorrefmark{1}Department of Electrical, Electronic and Information Engineering - University of Bologna, Italy\\
\IEEEauthorrefmark{2}Integrated System Laboratory - ETH Zurich, Switzerland\\
\IEEEauthorrefmark{3}GreenWaves Technologies, Grenoble, France\\
}
\vspace{-7mm}
}

\begin{document}

\maketitle

\begin{abstract}
In this paper, we present the first (to the best of our knowledge) demonstration of a low-power MCU-based edge device for Automatic License Plate Recognition (ALPR).
The design leverages on
% the flexibility and energy-efficiency of 
a  9-core RISC-V processor, GAP8, coupled with a QVGA ultra-low-power greyscale imager. 
The proposed visual processing pipeline uses a multi-model inference approach based on SSDlite-MobilenetV2 for license plate detection and LPRNet for optical character recognition, reaching a 38.9\% mAP score for the first task and a recognition rate of \textgreater99.13\% 
for the latter on public datasets.
On real-world data, the pipeline recognizes registration numbers when the size of LP crops is as small as 30$\times$5 pixels.
Thanks to the applied compression and optimization strategies, the multi-model inference (\SI{687}{\mega MAC}) achieves a throughput of \SI{1.09}{FPS} at a power cost of \SI{117}{\milli\watt} when running on GAP8. 
Our solution is the first MCU-class device embedding such a level of network complexity, resulting to be 73$\times$ more energy-efficient w.r.t. precedent mobile-class ALPR system featuring a Raspberry Pi3.
The proposed design does not resort to any hardwired acceleration engines, thus retaining full flexibility for future algorithmic improvements.

\end{abstract}

%%%%%%%%%%%%%%%%%%%%%%%%%%%%  Introduction %%%%%%%%%%%%%%%%%%%%%%%%%%%%%%%%%%%
\vspace{+5pt}

\section{Introduction }
Automatic License Plate Recognition (ALPR) is an optical character recognition (OCR) task consisting of detecting and transcribing the vehicle's License Plates (LPs).
Currently, Deep Learning (DL) algorithms are  the state-of-the-art (SoA) solution to this problem \cite{ccpd_rpnet}.
However, because of the high complexity (\textgreater\SI{10}{\giga MAC}) and stringent latency requirements, the deployment of ALPR pipelines is limited to high-performance embedded platforms that feature a high power envelope (\textgreater5 Watts) \cite{yolo_brazil,ssd_openvino,FPGA_1}: such systems would operate for no longer than 30 minutes when powered with a single AA battery, hence not suitable as an always-on smart sensing solution. 

Conversely, MicroController Units (MCUs) are the typical data processing engines for battery-powered systems, because of the low-cost and low-power characteristics (as low as few mW) and their SW programmability. However, MCUs, e.g. ARM Cortex-M4-based  devices, present severe limitations in terms of on-chip memory budget (few MB) and computing resources (a single-core running up to few hundreds of MHz) that prevent the implementation of complex vision DL pipelines~\cite{visual_wake_words}. 
Indeed, a typical SSD object detector is estimated to execute in \SI{3}{\second} at \SI{262}{\milli\watt} on an STM32H7 MCU~\cite{CMix-NN}, hence not meeting the requirements of low-power embedded devices.

\begin{comment} % stima di tempo necessaria per fare DL su un Cortex M7
STM32H7 ARM Cortex M7
image res: 192x192
network: MobilenetV1 x0.5 % YOLOv3-tiny	 5.56G ops
model size: 1.37MB
Latency: 510ms
MAC/Cycle: 0.40
energy: 134mJ % 262mW
--------------------------
My Object detector: 600M MAC
cycles: 1.5 G @0.4 MAC/cycle
freq: 480MHz
estimated inference time : 3.1 s     @262mW = 812mJ
\end{comment}

To tackle this challenge, in this work we propose a vision pipeline for ALPR, which relies on a 2-step DL-based algorithm composed of a SSD object detector~\cite{SSD} and LPRNet~\cite{LPRNet}, tailored for LP detection and recognition respectively. 
To gain an ultra-high energy-efficient smart sensing solution, the design of the DL model is optimized for the deployment on an ultra-low power smart camera platform, which includes GAP8 \cite{GAP8}, a Parallel Ultra-Low-Power (PULP) \cite{pulp-nn} architecture featuring an 8-core cluster that has a peak computing capability $\sim$\SI{1}{\giga MAC/\second} within $\sim$\SI{100}{\milli\watt} of power envelope, and an ultra-low-power Himax image sensor.
To fit the memory, latency, and energy consumption constraints of the target platform, we apply a multi-objective optimization procedure aiming at minimizing both the computational complexity (MAC operations) and the model size of a baseline solution~\cite{yolov3_embedded}, which exceed by far the capacity of deeply embedded devices.
In addition, the design process takes into account the reduced quality of the data produced by the low-power sensor, by ensuring high recognition capabilities of the proposed smart visual node.

The main contributions of this paper are:

\begin{enumerate}[]
    \item We present, for the first time,  an end-to-end solution, including system and algorithm design, for enabling ALPR on an energy-constrained MCU-based visual edge-node;

    \item  We optimize and compress the detection and recognition stages of the ALPR pipeline for efficient deployment on the targeted MCU-based platform;
    
    \item We provide an experimental analysis on both real-world data and publicity-available datasets in terms of accuracy, inference speed, and energy efficiency and we compare our ALPR solution w.r.t other SoA systems.
    
\end{enumerate}

The proposed design leads to the first smart sensor device embedding a visual pipeline for ALPR within a full-system power envelope of \SI{117}{\milli\watt}\footnote{The application code is open-source at:
\url{https://github.com/GreenWaves-Technologies/licence_plate_recognition}}, which accounts the sensor, memories, and processor power costs.
The effectiveness of our design is demonstrated by testing our system on public benchmark datasets and real-world data, obtaining a mAP score of 38.9\% for license plate detection and an accuracy of \textgreater99.13\% for character recognition. Our system shows a processing throughput \SI{1.09}{FPS} at an energy cost of \SI{108}{\milli\joule} and results 73$\times$ more energy-efficient w.r.t. precedent mobile-class ALPR system featuring a Raspberry Pi3 (ARM Cortex-A53 CPU) coupled to a Pi NoIR camera \cite{yolov3_embedded}.

%%%%%%%%%%%%%%%%%%%%%%%%%%%%%%%%%%%  Related Works  %%%%%%%%%%%%%%%%%%%%%%%%%%%%%%%%%%%%%
\begin{figure*}[h]
  \centering
  \includegraphics[width=\linewidth]{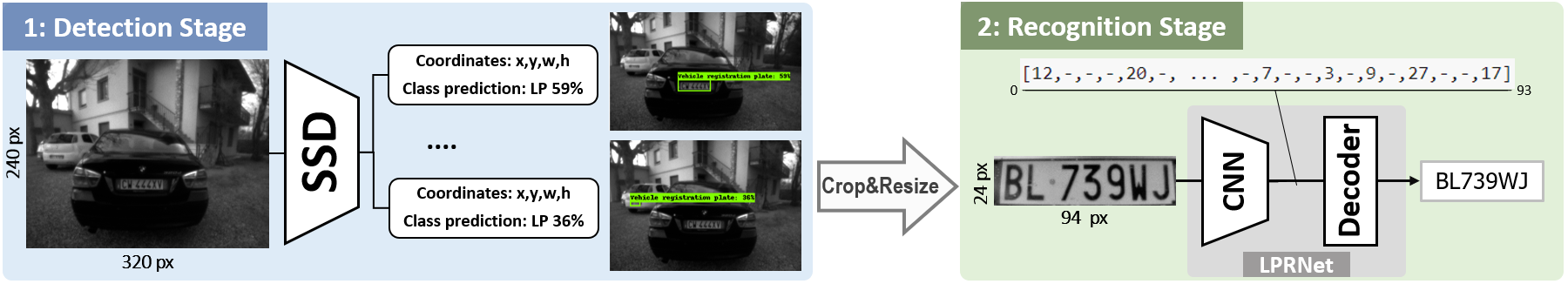}
%   \vspace{-1mm}
  \caption{The two-stage license plate detection and recognition pipeline proposed. The input image is first processed by SSD, which detects all the license plates in an image, then it is processed by LPRNet, which recognize the registration numbers.}
  \label{fig:SSD_and_LPRNet}
  \vspace{-3mm}
\end{figure*}

\section{Related Works}
In this section we review the OCR algorithms for ALPR and the low-power embedded solutions that address this task.

\textbf{OCR algorithms for ALPR.}
This task is typically addressed as a two-step process: a detection stage estimates the plates' locations and a recognition step outputs the transcription of the characters.
Single-shot object detectors represent a popular solution for LP detection. YOLO \cite{YoloV3} was deployed in \cite{yolo_a_two_stage_deep_neural_network, yolo_brazil}, but these works maximize accuracy leveraging on high complexity models (\textgreater\SI{10}{\giga MAC}) and making use of  powerful GPUs for the inference task.  To target embedded implementations, \cite{ssd_openvino} uses SSD \cite{SSD} coupled to a custom lightweight CNN architecture ($\sim$3M parameters).
Likewise, we start from an SSD object detector featuring a MobilenetV2 backbone for LP detection and we optimize it for the deployment on the target system. 
Concerning the character recognition task, segmentation-based methods make use of object detectors to detect individual characters \cite{yolo_brazil, ssd_openvino}. This approach is prone to errors since a single character miss-detection leads to a wrong recognition \cite{segmentation_problem}.
In contrast, segmentation-free approaches produce high-accurate readings  \cite{what_is_wrong_with_text_recognition}, outrunning segmentation-based approaches by $5\times$ when fed with the LP crops \cite{yolo_a_two_stage_deep_neural_network}. 
Among them, LPRNet~\cite{LPRNet} is a fully convolutional model that features 1.5M parameters, provides high accuracy and runs at \SI{333}{FPS} on an \textit{Intel Core i7-6700K CPU}. In this work, we select this model as a baseline for LP character recognition: its fully-convolutional nature is features a higher efficiency on
the targeted HW/SW platform w.r.t. RNN-based memory-bounded solutions~\cite{RNNs_embedded}.

\textbf{Energy-efficient OCR systems.}
In recent years, several embedded systems for ALPR have been developed, but they are not intended to target ultra-low-power OCR applications on a tiny form factor device. 
The solutions~\cite{FPGA_1} and \cite{FPGA_2} leverage Xilinx Zynq FPGAs to implement LP detection and character recognition respectively: both systems run in real-time, but with a power consumption of \SI{6.9}{\watt} and \SI{8}{\watt} respectively. \textit{Izidio} \cite{yolov3_embedded} performs both steps making use of a Raspberry Pi device coupled to a Pi NoIR V2 camera featuring a power consumption of \SI{3.12}{\watt}. In contrast, on the low-end side, Pixy2 \cite{pixy2} is a smart camera system that can perform object detection tasks, but still with a typical power consumption of \SI{700}{\milli\watt}. To improve  performance, NN accelerators can be used. \textit{Castro-Zunti et al.}~\cite{ssd_openvino} exploits an Intel Neural Compute Stick 2 (NCS2) to perform LP detection and recognition, processing $300\times300$ images in real-time at \SI{15}{FPS}. Still, the power required by the NSC2 ranges between \SI{1.5}{\watt} and \SI{2}{\watt}. 
To reduce the power consumption, Kendryte K210 exploits a hardwired NN accelerator, but when integrated into a general-purpose AI vision system, e.g. HUSKYLENS \cite{kendryte}, it features a power consumption of \SI{700}{\milli\watt}.\\
Ultra-low-power MCUs fit into the targeted power envelope ($\sim$\SI{100}{\milli\watt}) but, because of their limited computing power, no complex and real-time DL-based applications have been demonstrated  \cite{keyword_spotting,visual_wake_words, CMix-NN}.
Related to our work, Palossi et al.~\cite{palossi_nano_drone} deployed an NN-based complex autonomous navigation task on a nano-sized drone by exploiting the parallelism of the GAP8 processor under a power envelope of \SI{64}{\milli\watt}.
In contrast, in this paper we show the embedding process of a more complex 2-steps DL algorithm into the PULP-based GAP8 platform, demonstrating the first implementation of a full visual OCR pipeline for LP detection and recognition on such a tiny system.

%%%%%%%%%%%%%%%%%%%%%%%%%%%%%%%%%%%%  Visual System  %%%%%%%%%%%%%%%%%%%%%%%%%%%%%%%%%%%%%

\section{MCU-based Visual System for ALPR}
In this section, we describe the components of the proposed visual system, including both the HW/SW building blocks and the 2-steps computer vision pipeline for ALPR (Fig.~\ref{fig:SSD_and_LPRNet}).
%, which is composed by a Single-Shot Detector (SSD) for LP detection and LPRNet for LP recognition.

\subsection{GAP8-based vision system}
% gap8
The smart vision system is based on the GAP8~\cite{GAP8} processor, an embedded RISC-V 9-core processor derived from the PULP open-source project \cite{pulp-nn}. GAP8 includes a RISC-V single-core sub-system enriched with a full-set of peripherals and a \SI{512}{\kilo \byte} L2 memory. In addition, an 8-cores cluster featuring \SI{64}{\kilo \byte} of shared L1 memory is turned on on-demand to process compute-intensive workload, e.g. CNN inferences.
% rest of the HW
The processor is coupled with off-chip L3 memories, either a \SI{64}{\mega \byte} FLASH memory, for static data storage, and a \SI{8}{\mega \byte} of RAM memory, for dynamic data allocation, and interfaced to a low-power Himax HM01B0 image sensor, burning as low as few mW when sampling greyscale QVGA images ($320\times240 px$). 
% SW stack

To efficiently run DL inference tasks on the parallel cluster, the GAP\textit{flow} SW toolset can be fed with a network graph file (e.g., \texttt{tflite}) to produce a GAP-optimized C code. This latter implements a dataflow that distributes the workload, e.g. convolutional layer kernels, over the 8 cores of the cluster while transferring data from L2/L3 to L1 and vice versa in the background. Hence, the toolset produces computation parallel primitives that are optimized to run inference on 8-bit quantized DL models, by also exploiting DSP-oriented SIMD extensions. 
% Note that, given the signed arithmetic nature, the SW computation kernels are tailored for symmetrically quantized networks.

\subsection{ALPR Pipeline}
The proposed ALPR system operates in two steps: a plate detection stage and a plate recognition stage (Fig.\ref{fig:SSD_and_LPRNet}). 

\textbf{License Plate Detection.}
We address this task using the single-shot object detector SSD based on the MobilenetV2 architecture, which provides a good trade-off between speed and accuracy \cite{SSD}. This module takes raw images and produces bounding box coordinates for each LP in the image, as shown in Fig. \ref{fig:SSD_and_LPRNet}.
To improve performance, we replace the SSD heads with the SSDLite counterpart: the bounding box predictor layers are replaced by depth-wise separable convolution blocks, saving 1.5M parameters and reducing the computational complexity by \SI{30}{\mega MAC}~operations.

\textbf{License Plate Recognition.}
The character recognition task is addressed by using LPRNet \cite{LPRNet}, a fully-convolutional DL based segmentation-free approach.
This is composed of a CNN feature extractor and a decoder (Fig.\ref{fig:SSD_and_LPRNet}) for a total of only  1.7M parameters. 
The CNN is fed with a $94\times 24 px$ license plate crop and outputs a $m\times n$ 2D-tensors where the class probability concerning the $m$ characters of the vocabulary (including "-", the no-character symbol) is predicted for all the $n$ spatial segments (94 in our case).
The decoding stage makes use of a CTC greedy decoder to predict the vehicle registration number: for any spatial segment, the decoder selects the characters with the highest score and removes blank spaces and redundant predictions, if any.

% TABELLA SSD
\begin{table}[tb]
\footnotesize
\setlength{\tabcolsep}{0.41em}
\renewcommand{\arraystretch}{1.2}% for the vertical padding of tables
\centering
\caption{SSD optimization for LP detection: effects of architecture type and hyperparameters on the mAP score, MACs, and model size.}
\label{tab:ssd_results}
% \vspace{-2mm}
\begin{tabular}{lcccccc} 
\toprule
\multirow{3}{*}{}                                                             & \multirow{3}{*}{\begin{tabular}[c]{@{}c@{}}\textbf{Variable}\\\textbf{parameter} \end{tabular}} & \multicolumn{2}{c}{\begin{tabular}[c]{@{}c@{}} \textbf{mAP score}\\ \end{tabular}} & \multirow{3}{*}{\begin{tabular}[c]{@{}c@{}}\textbf{N°}\\\textbf{param} \end{tabular}} & \multirow{3}{*}{\begin{tabular}[c]{@{}c@{}} \textbf{Model}\\\textbf{Size} \\{[}MB] \end{tabular}} & \multirow{3}{*}{\begin{tabular}[c]{@{}c@{}} \textbf{MAC}\\\textbf{ops} \end{tabular}}  \\ 
\cline{3-4}
                                                                              &                                                                                                 & Open-Images        & Himax                                                                  &                                                                                       &                                                                                                   &                                                                                        \\
                                                                              &                                                                                                 & (1113 img)        & (63 img)                                                               &                                                                                       &                                                                                                   &                                                                                        \\ 
\midrule
\multirow{3}{*}{\begin{tabular}[c]{@{}l@{}}Network \\ topology \end{tabular}} & SSD-MV1                                                                                         & 39.6\%            & 58.7\%                                                                 & 5.5M                                                                                  & 22.0                                                                                              & 931M                                                                                   \\
                                                                              & SSD-MV2                                                                                         & 42.0\%            & 60\%                                                                   & 4.6M                                                                                  & 18.4                                                                                              & 540M                                                                                   \\
                                                                              & \textbf{SSDlite-MV2}                                                                            & 39.8\%            & 52.4\%                                                                 & 3.1M                                                                                  & 12.4                                                                                              & 515M                                                                                   \\ 
\midrule
\multirow{3}{*}{\begin{tabular}[c]{@{}l@{}}Image~\\size\end{tabular}}         & \textbf{320x240}                                                                                & 42.0\%            & 60\%                                                                   & 4.6M                                                                                  & 18.4                                                                                              & 540M                                                                                   \\
                                                                              & 352x352                                                                                         & 42.1\%            & 60\%                                                                   & 4.6M                                                                                  & 18.4                                                                                              & 840M                                                                                   \\
                                                                              & 512x512                                                                                         & 45.4\%            & 57.3\%                                                                 & 4.6M                                                                                  & 18.4                                                                                              & 1.76G                                                                                  \\ 
\midrule
Quant                                                                      & \textbf{8-bit}                                                                                  & \textbf{38.9\% }  & \textbf{49.4\% }                                                       & \textbf{3.1M}                                                                         & \textbf{3.1}                                                                                      & \textbf{515M}                                                                          \\ 
\midrule
\multirow{2}{*}{\begin{tabular}[c]{@{}l@{}}Training\\samples \end{tabular}}   & \textbf{RGB}                                                                                    & 42.0\%              & 60\%                                                                   & 4.6M                                                                                  & 18.4                                                                                              & 540M                                                                                   \\
                                                                              & Greyscale                                                                                       & 39.7\%            & 56.4\%                                                                 & 4.6M                                                                                  & 18.4                                                                                              & 540M                                                                                   \\
\bottomrule
\end{tabular}
\vspace{-3mm}
\end{table}

\section{ALPR pipeline optimization}
\label{sec:optimization}

In this section, we showcase the multi-criteria  optimization process for an efficient embedding of the 2-steps ALPR pipeline into an MCU device. The design aims at reducing both the computational complexity (MAC operations) and the model size at a low accuracy penalty w.r.t. the baseline.

\squeezeup

\subsection{Optimization Environment}

\textbf{Evaluation Metrics.} The LP detection efficacy is measured by computing the \textit{mean Average Precision (mAP)}\cite{COCOmAP}.
On the other side, the character recognition task is evaluated with the \textit{License Plate Recognition Rate} (LP-RR) metric, which accounts the number of correctly recognized license plates over the total number of license plates. 

\textbf{Datasets.}
The LP detection model is trained on a dataset distilled from OpenImagesV4~\cite{dataset_OpenImages} by including images of LPs and negative examples.
The recognition algorithm is evaluated on 3 LP datasets that differ for country, color, and layout of the plates: CCPD (Chinese City Parking Dataset) \cite{ccpd_rpnet}, Synthetic Chinese LPs\cite{dataset_SynthChinese}, and ReId Czech LPs\cite{dataset_Czech}.
The license's plate nationality impacts the number of classes: Chinese plates have a total of 70 characters classes making it a harder problem w.r.t. ReId Czech plates, which consists only of 36 classes. To assess the model's performance on a real setup, we collected the \textit{Himax Test Dataset}, composed of 48 Chinese and 15 Italian license plates acquired with the Himax camera. The dataset includes multiple snapshots (QVGA resolution) of 15 vehicles  with an increasing distance from the vehicle: 0.3m, 1m, 2m, 3m (Fig.~\ref{fig:recognition_distances}): the LPs feature a varying pixel size that ranges from $200\times50 px$ to $30\times5 px$.

\subsection{License Plate Detection stage}
\label{sec:experimental_results_detection}

For each optimization step, Tab.~\ref{tab:ssd_results}  
reports the mAP score, the number of parameters, the model size, and the number of MACs on two testing datasets.\\
\textbf{Network Topology.} 
Our baseline SSD-MobilenetV2 achieves a 42\% mAP for license plate detection on OpenImagesV4 ($320\times240 px$ input), +2.4\% higher than SSD-MobilenetV1. 
However, we choose the SSDLite-MobilenetV2 model as the final architecture, which achieves a mAP score of 39.8\% but features a much lower memory footprint (3.08M parameters) and a lower \SI{515}{\mega MAC} computational cost (thanks to depthwise convolutional blocks), leading to a faster inference time. \\
\textbf{Input size.} SSD-MobilenetV2 scores 42.0\% mAP when trained and tested on images resized to $320\times240$ px (i.e., the Himax camera resolution). Increasing the resolution improves the detection accuracy, but the MAC cost increases exponentially.
Hence, we choose a model input size of $320\times240$ px, which leads to the solution with the lowest computational cost (\SI{540}{\mega MAC}) and shows to be more effective on the images collected with our Himax sensor.\\  
\textbf{Training on Grayscale Samples.} 
Because our Himax image sensor produces greyscale images, we convert the samples from RGB to greyscale before feeding the SSD-MobilenetV2 model at training time. 
On the Himax Test dataset (63 samples), we observe a higher mAP accuracy when the model was trained on RGB images (+3.6\%), suggesting that RGB format leads to a higher generalization when inferencing on greyscale data. \\
\textbf{Quantization.} We make use of quantization aware training \cite{quantization_jacob} to compress the network from 32-bit to 8-bit precision, reducing the memory footprint by $4\times$. 
The final SSDlite-MobilenetV2 quantized model only loses 0.9\% mAP w.r.t. the full-precision model, scoring a 38.9\% mAP on the OpenImagesV4~dataset and 49.4\% mAP on the Himax Test dataset.

This model is $17\times$ more efficient in terms of MAC operations to the smallest object detection counterpart YOLO Nano \cite{yolo_nano}, which features a \SI{4}{\mega \byte} model size with \SI{9.14}{\giga MAC} operations.

% Table LPRNet results
\begin{table}[tb]
% \vspace{-2mm}
\footnotesize
\setlength{\tabcolsep}{0.53em}
\centering
\caption{LPRNet optimization for LP recognition: effects on the MACs, model size, and LP Recognition Rate (LP-RR) over 3 datasets.}
% \vspace{-2mm}
\label{tab:lprnet_results}
\begin{tabular}{llcccr} 
\toprule
\multirow{2}{*}{\rule{0pt}{2.2ex}   \textbf{LPRNet} } & \multicolumn{1}{c}{\multirow{2}{*}{\begin{tabular}[c]{@{}c@{}} \textbf{N°params}\\\textbf{(size)} \end{tabular}}} & \multicolumn{1}{c}{\multirow{2}{*}{\begin{tabular}[c]{@{}c@{}} \textbf{MAC}\\\textbf{ops} \end{tabular}}} & \multicolumn{3}{c}{\textbf{dataset LP-RR} }                                          \\ 
\cline{4-6}
\rule{0pt}{2.2ex}  
                                   & \multicolumn{1}{c}{}                                                                                              & \multicolumn{1}{c}{}                                                                                      & \multicolumn{1}{c}{CCPD} & \multicolumn{1}{c}{Synthetic} & \multicolumn{1}{c}{ReId}  \\ 
\midrule
Baseline Net.                      & 1.7M (6.8MB)                                                                                                      & 172M                                                                                                      & 99.59\%                  & 98.14\%                       & 99.52\%                   \\
Layer replaced                     & 2.4M (9.6MB)                                                                                                      & 173M                                                                                                      & 99.53\%                  & 97.50\%                        & 99.40\%                   \\
FC Layer halved                    & 1.0M (4MB)                                                                                                        & 172M                                                                                                      & 99.30\%                  & 97.70\%                        & 99.22\%                   \\
8-bit Quantization                 & 1.0M (1MB)                                                                                                        & 172M                                                                                                      & 99.13\%                  & 97.66\%                       & 99.22\%                   \\
\bottomrule
\end{tabular}
\vspace{-4mm}
\end{table}

\subsection{License Plate Recognition stage}

License plate crops are resized to a fixed $94\times 24 px$ size before feeding the recognition model. Models are trained from scratch  with random initialization of the weight parameters.
Tab.~\ref{tab:lprnet_results} reports the optimization results.
\\
\textbf{Network Topology.} The original LPRNet topology~\cite{LPRNet} is refined for two reasons: remove unsupported operations by the GAP\textit{flow} and reduce the memory footprint (baseline version features 1.7M parameters). 
In the original network, the output of the CNN features extractor is augmented with the global context embedding as in~\cite{ParseNet}: the backbone output is processed with a fully connected layer, tiled (i.e., replicated), and then it is concatenated to the backbone output itself. 
Since this tiling operation is not yet supported by the GAP\textit{flow},
this layer has been replaced with a fully connected layer featuring the desired output size. 
This adaptation  has a negative effect on both memory footprint (increased by  $1.4\times$) and LP-RR accuracy (decreased by 0.1-0.6\%). To mitigate the growth of the model's size, we half the depth of the first fully connected layer after the backbone's output: this reduces the memory footprint of \SI{5.6}{\mega \byte} while keeping almost the same accuracy levels w.r.t. the full-size network.\\
\textbf{Quantization.} The model is compressed to 8-bit by means of symmetric quantization. We use quantization-aware training to mitigate the accuracy degradation. Thanks to this, the network's size is reduced from \SI{4}{\mega \byte} to \SI{1}{\mega \byte} while degrading the LP-RR metric by just 0.04-0.17\%  (Tab. \ref{tab:lprnet_results}). The final scores are 99.13\%, 97.66\%, and 99.22\% on CCPD, Synthetic Chinese LPs, and ReId Czech LPs datasets respectively.\\

\begin{table}[t]
\footnotesize
\setlength{\tabcolsep}{0.5em}
\centering
\caption{Summary of memory footprint and speed performances of the final OCR pipeline composed of 2 stages: SSD for LP detection (D) and LPRNet for LP recognition (R).}
\label{tab:conclusion}
% \vspace{-2mm}
\begin{tabular}{ccccccc} 
\toprule
\textbf{Algorithm}  & \textbf{Task}  & \begin{tabular}[c]{@{}c@{}}\textbf{N°} \\\textbf{params} \end{tabular} & \begin{tabular}[c]{@{}c@{}} \textbf{MAC}\\ \textbf{ops} \end{tabular} & \textbf{Cycles}  & \begin{tabular}[c]{@{}c@{}}\textbf{Proc time}\\ @175MHz \end{tabular} & \begin{tabular}[c]{@{}c@{}}\textbf{FPS}\\ @175MHz \end{tabular}  \\ 
\midrule
SSD                 & D              & 3.1M                                                                   & 515M                                                                  & 101M             & 0.577s                                                                & 1.73 FPS                                                         \\
LPRNet              & R              & 1.0M                                                                   & 172M                                                                  & 60M              & 0.343s                                                                & 1.62 FPS                                                         \\
SSD+LPRNet          & D+R            & 4.1M                                                                   & 687M                                                                  & 161M             & 0.92s                                                                 & 1.09 FPS                                                         \\
\bottomrule
\end{tabular}
\vspace{-3mm}
\end{table}

\begin{table}[b]
\vspace{-2mm}
\footnotesize
\setlength{\tabcolsep}{0.75em}
\centering
\caption{ALPR pipeline tested on minimum LP size for detection and recognition (setup in Fig.\ref{fig:recognition_distances}) over Himax Test dataset (63 images). The maximum LP distance depends on the focal length of the camera.}
\label{tab:pipeline_vs_distance}
\begin{tabular}{lcccc}
\toprule 
 \textbf{LP size}  & \textbf{LP detected}  & \textbf{Char size}  & \textbf{LP-RR}  & \textbf{Distance}  \\ 
\midrule
200x50px           & 15/15                 & 20x40 px            & 100\%           & 0.3m      \\ 

100x20px           & 15/15                 & 8x16 px             & 100\%           & 1m        \\ 

50x10 px           & 14/15                 & 5x10 px             & 36.4\%          & 2m        \\ 

40x7  px           & 15/15                 & 3x7 px              & 0\%             & 3m        \\ 

30x5  px           & 3/3                   & 2x5 px              & 0\%             & 4m        \\
\bottomrule
\end{tabular}
\vspace{-2mm}
\end{table}

\squeezeup

\section{Experimental Results}
In this section, we provide experiments on in-field data and we give details about the energy efficiency of our system.

\begin{figure}[tb]
  \centering
  \includegraphics[width=\linewidth]{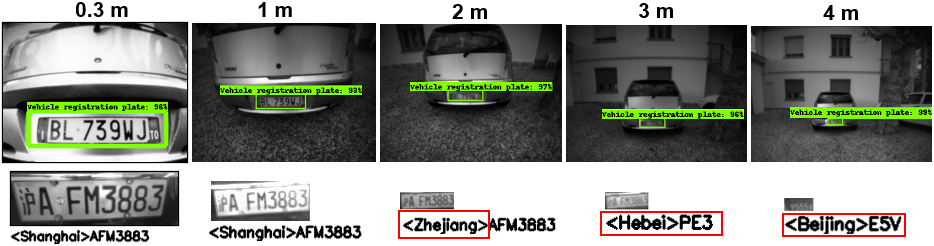}
  \caption{Inference on the Himax dataset. The maximum LP detection distance is 4m, while the maximum recognition distance is 1m.}
  \label{fig:recognition_distances}
\end{figure}

\textbf{End-to-end ALPR pipeline evaluation}.
Tab. \ref{tab:conclusion} summarizes the model size and complexity of the designed  2-stage ALPR pipeline.
A total of 4.1M parameters is accounted, compliant with the memory characteristics of GAP8: 3.1M parameters for the SSD detector and 1.0M parameters for LPRNet. 
We validated the optimized 8-bit ALPR models on the real-world data provided by the Himax Test dataset (63 images). 
Tab.~\ref{tab:pipeline_vs_distance} reports the detection and the recognition accuracy when varying the distance of the vehicle from the image sensor, while Fig.~\ref{fig:recognition_distances} illustrates the experimental setup. 
On one side, the detection model achieves high accuracy in the full range, even when the LP crop gets as small as $30\times 5$ pixels: only one LP is miss-detected over the 63 images. 
On the other hand, the character recognition model shows a $100\%$ LP-RR with a minimum character size of $8\times 16$ pixels. Then the recognition rate dramatically decreases under $36.4\%$ 
when the characters feature a spatial size smaller than  $5\times 10$ pixels, becoming unreadable. 
In our setup, the maximum recognition distance with $100\%$ recognition rate is found with LPs at a distance of 1m. Note that this distance is heavily influenced by the focal length of the camera's lens, which can be tuned to the application scenario.

% Tabella energy efficiency
\begin{table}[tb]
\centering
\footnotesize
\setlength{\tabcolsep}{0.5em}
\caption{Energy comparison with existing ALPR embedded systems. The tasks addressed are categorized in: LP detection (D), LP Recognition (R), Camera acquisition of images (C).}
\label{tab:energy_comparison}
\begin{tabular}{lcccccc} 
\toprule
 \begin{tabular}[c]{@{}l@{}}\textbf{ALPR }\\\textbf{System}\end{tabular}     & \textbf{Platform } & \textbf{Power } & \begin{tabular}[c]{@{}c@{}}\textbf{Infer }\\\textbf{ time }\end{tabular} & \begin{tabular}[c]{@{}c@{}}\textbf{Energy }\\\textbf{[mJ] }\end{tabular} & \textbf{MHz }  & \textbf{Tasks }  \\ 
\midrule
\textbf{Our work}  & \textbf{GAP8}      & \textbf{117mW}  & \textbf{0.92s}                                                               & \textbf{108 }                                                            & \multicolumn{1}{l}{\textbf{ 175} }  & \textbf{C,D,R}   \\
\textit{Izidio} \cite{yolov3_embedded}    & Raspberry Pi3      & 3.12W           & 2.70s                                                                         & 8420                                                                     & -              & C,D,R            \\
\textit{Li} \cite{FPGA_1}       & FPGA               & 6.9W            & 55ms                                                                         & 380                                                                      & 215            & D                \\
\textit{Fan}  \cite{custom_fpga_acc}      & FPGA               & 9.9W            & 15ms                                                                         & 150                                                                      & 100            & D                \\
\textit{Rybalkin}\hspace{1sp}\cite{FPGA_2}  & FPGA               & 7.0W            & 1.17s                                                                        & 8150                                                                     & 142            & R                \\
\textit{Castro}\cite{ssd_openvino}     & INTEL NCS2         & 2.0W            & 66ms                                                                         & 132                                                                      & 600            & R                \\
\bottomrule
\end{tabular}
\vspace{-4mm}
\end{table}

%%%%%%%%%%%%%%%%%%%%%%%%%%%%%%%%%%%%%%%%%%%%%%%%%%%%%%%%%%%%%%%%%%%%

\textbf{Energy evaluation and comparison.}
Tab.~\ref{tab:conclusion} summarizes the performance  of the ALPR pipeline. 
Despite our compressed model can be mapped to any MCU architecture, we exploit the PULP \cite{pulp-nn} architecture of GAP8 to gain fast inference time and low energy over typical single-core MCUs.
The combination of the detection stage (\SI{515}{\mega MAC}) and the recognition stage (\SI{172}{\mega MAC}) features a total latency of 161M clock cycles on GAP8 (8-cores cluster @ \SI{175}{\mega \hertz}), by leveraging on the Gap\textit{flow} toolset. This results in an inference time of \SI{0.92}{\second} for the whole pipeline, corresponding to a throughput of \SI{1.09}{FPS}.
During the inference process, GAP8 features an average power consumption of \SI{108}{\milli\watt}. The external L3 memory adds \SI{8}{\milli\watt} to the system power cost, while the Himax camera consumes as little as \SI{2}{\milli\watt} when sampling at the lowest frame-rate (\SI{30}{FPS}). The system-level average power consumption is \SI{117}{\milli\watt} and the total inference energy cost~is~\SI{108}{\milli\joule}.

Tab. \ref{tab:energy_comparison} compares our solution against other publicly available ALPR implementations. 
\textit{Izidio et al.}~\cite{yolov3_embedded} relies on a general-purpose processor and a camera sensor featuring an energy consumption of \SI{8.42}{\joule} for inference (based on YOLOv3 \cite{YoloV3} object detector), being  $ 73\times$ more energy-demanding than our system. % and $3\times slower$.
Other related works~\cite{FPGA_1,custom_fpga_acc,FPGA_2} reported in Tab. \ref{tab:energy_comparison} address only a part of the ALPR problem, either detection or recognition, and they do not consider image acquisition tasks or data quality. Still, they feature a higher energy consumption per inference w.r.t. our work.

\section{Conclusion}

We presented an MCU-based system, which includes a 9-core IoT processor, GAP8, and a low-power QVGA imager, for license plate detection and recognition.  
The proposed vision pipeline achieved a mAP score of 38.9\% on the OpenImagesV4 dataset (license plates only) for the detection task and a LP recognition rate of \textgreater$99.13\%$ on multiple LPs datasets. 
The optimized multi-model visual pipeline features 4.1M parameters and \SI{687}{\mega MAC}, turning into an inference time of 0.92s on GAP8 running at \SI{175}{\mega \hertz}. 
Our solution results $ 73\times$ less energy demanding than previous general-purpose systems, based on Raspberry Pi, and demonstrated, for the first time, such a level of network-complexity on an MCU-based device.

\bibliographystyle{IEEEtran}
\bibliography{bibliography}
% \bibliography{IEEEabrv,bibliography}

\end{document}